\documentclass[sigconf]{acmart}

\AtBeginDocument{%
  \providecommand\BibTeX{{%
    \normalfont B\kern-0.5em{\scshape i\kern-0.25em b}\kern-0.8em\TeX}}}

\copyrightyear{2019}
\acmYear{2018}
\setcopyright{acmlicensed}
\acmConference[MileTS '19]{MileTS '19: 5th KDD Workshop on Mining and Learning from Time Series}{August 5th, 2019}{Anchorage, Alaska, USA}
\acmBooktitle{MileTS '19: 5th KDD Workshop on Mining and Learning from Time Series, August 5th, 2019, Anchorage, Alaska, USA}
\acmPrice{15.00}
\acmDOI{10.1145/1122445.1122456}
\acmISBN{978-1-4503-9999-9/18/06}

\usepackage{array}
\usepackage{multirow}
\usepackage{amsmath}
\usepackage{url}
\usepackage{amsfonts}
\usepackage{bbm}
\usepackage[noend, noline, algoruled, linesnumbered]{algorithm2e}

\usepackage{pgf}
\usepackage{hyperref}
\newcommand{\xhdr}[1]{{\bf #1.}}
\newcommand{\quotes}[1]{``#1''}
\usepackage{tikz}
\usetikzlibrary{arrows,automata}
\usepackage{tikz}
\newcommand{\bbS}{\mathbf{S}}
\usepackage{subfigure}
\usepackage{mathrsfs}  
\SetKwProg{Fn}{Function}{}{}

\begin{document}

%
% The "title" command has an optional parameter, allowing the author to define a "short title" to be used in page headers.
\title[MASA: Motif-Aware State Assignment]{MASA: Motif-Aware State Assignment in Noisy Time Series Data}

%
% The "author" command and its associated commands are used to define the authors and their affiliations.
% Of note is the shared affiliation of the first two authors, and the "authornote" and "authornotemark" commands
% used to denote shared contribution to the research.
\author{Saachi Jain}
\email{saachi@stanford.edu}
\affiliation{Stanford University}
\author{David Hallac}
\email{hallac@stanford.edu}
\affiliation{Stanford University}
\author{Rok Sosic}
\email{rok@stanford.edu}
\affiliation{Stanford University}
\author{Jure Leskovec}
\email{jure@stanford.edu}
\affiliation{Stanford University}

% \affiliation{%
%   \institution{Stanford University}
% }

%
% By default, the full list of authors will be used in the page headers. Often, this list is too long, and will overlap
% other information printed in the page headers. This command allows the author to define a more concise list
% of authors' names for this purpose.
\renewcommand{\shortauthors}{Jain, et al.}

%
% The abstract is a short summary of the work to be presented in the article.
\begin{abstract}
  Complex systems, such as airplanes, cars, or financial markets, produce multivariate time series data consisting of a large number of system measurements over a period of time. Such data can be interpreted as a sequence of \emph{states}, where each state represents a prototype of system behavior.  An important problem in this domain is to identify repeated sequences of states, known as \emph{motifs}. Such motifs correspond to complex behaviors that capture common sequences of state transitions.
  For example, in automotive data, a motif of ``making a turn'' might manifest as a sequence of states: slowing down, turning the wheel, and then speeding back up. However, discovering these motifs is challenging, because the individual states and state assignments are unknown, have different durations, and need to be jointly learned from the noisy time series. Here we develop \emph{motif-aware state assignment} (MASA), a method to discover common motifs in noisy time series data and leverage those motifs to more robustly assign states to measurements. We formulate the problem of motif discovery as a large optimization problem, which we solve using an expectation-maximization type approach. MASA performs well in the presence of noise in the input data and is scalable to very large datasets. Experiments on synthetic data show that MASA outperforms state-of-the-art baselines by up to 38.2\%, and two case studies demonstrate how our approach discovers insightful motifs in the presence of noise in real-world time series data.
  \end{abstract}


%
% Keywords. The author(s) should pick words that accurately describe the work being
% presented. Separate the keywords with commas.

%
% This command processes the author and affiliation and title information and builds
% the first part of the formatted document.
\maketitle
\section{Introduction}

Many domains and applications, ranging from automobiles~\cite{TICC}, financial markets~\cite{fu2001evolutionary}, and wearable sensors \cite{vahdatpour2009toward}, generate large amounts of time series data. In most cases, this data is multivariate, where each timestamped measurement consists of a vector of readings from multiple entities, such as sensors. Thus, multivariate time series data captures system state via a sequence of sensor readings.

However, directly extracting meaningful insights from time series data is challenging, since the relationships between readings are often complex and mutated by noise. Not all sensors are important all the time, and key information may lie in the connection between readings across different timestamps rather than within individual measurements themselves. To understand this complex data, it is useful to label each measurement as one of $K$ unique states. Each state is an interpretable template for system behavior which can repeat itself many times across the time series. These states distill the complexities of the multivariate dataset into a more accessible and interpretable symbolic representation. Moreover, this task allows us to partition the measurements in the time series into variable-length segments, where each segment is a sequence of measurements in the same state. Segmentation allows one to characterize large windows of time that exhibit a single behavior.

\begin{figure}[t!]
    \includegraphics[width=0.5\textwidth]{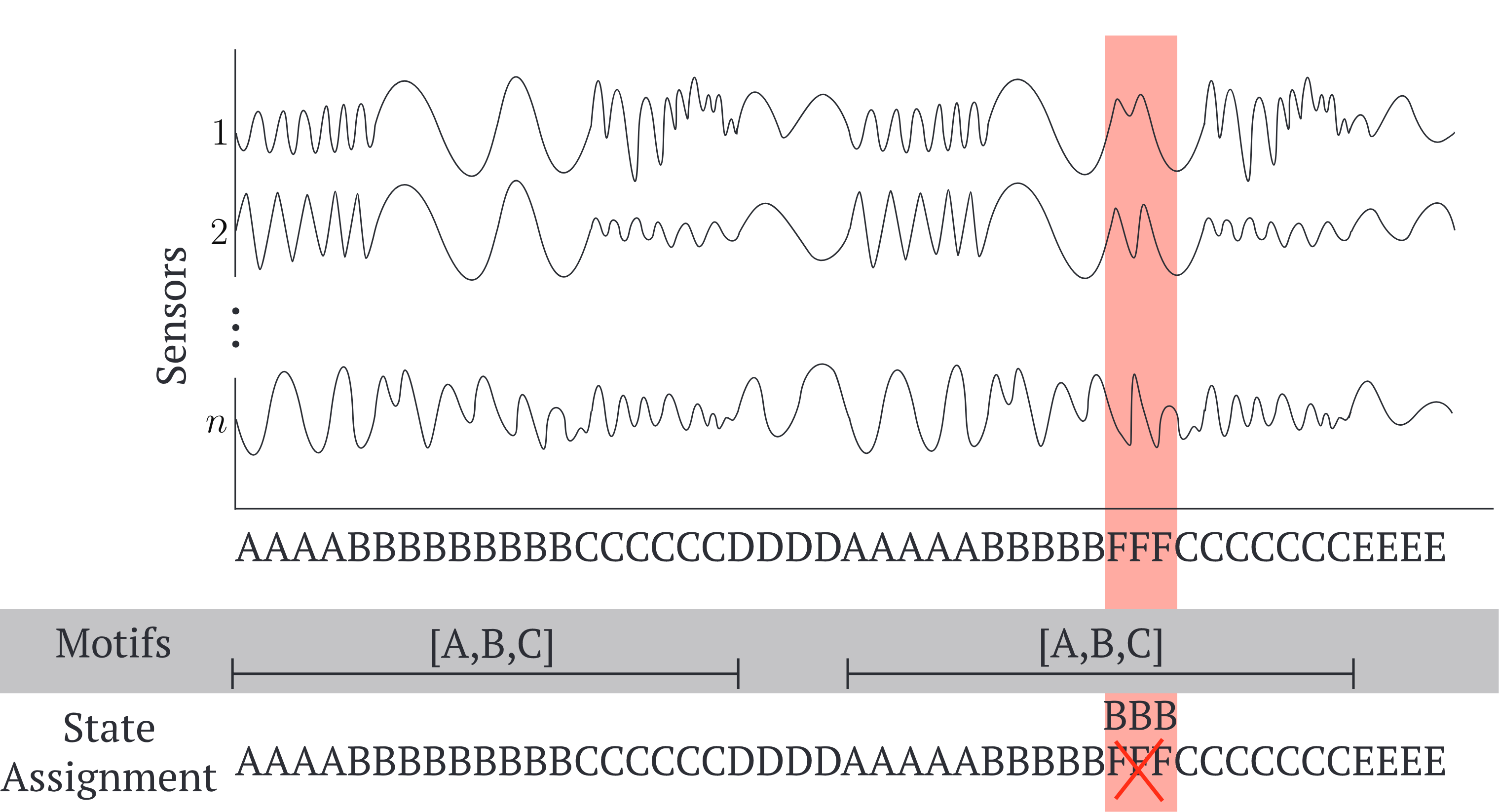}
	\vspace{-6mm}
    \caption{
    MASA assigns each measurement in a multivariate time series to a state. MASA further discovers motifs (repeated sequences of states), using them to improve state assignment in the presence of noise. Furthermore, the same motif ($[A,B,C]$) can spend different amount of time in each state (4 vs. 5 measurements in state $A$).
    }
	\vspace{-6mm}
    \label{Fig1}
\end{figure}

These states do not appear in isolation; the context in which a state occurs provides critical insights and can be just as useful as the state itself. It is therefore important to discover \emph{motifs}, which represent a sequence of state transitions. For example, in a car, a single state may indicate that the driver is slowing down, but the sequence of slowing down, turning the wheel, and then speeding up implies that the vehicle is in a motif known as a ``turn''. A motif is then an abstraction over a sequence of states in the time series. Motifs specify transitions but do not constrain duration: because there are many types of turns, a turning motif must not specify how long a driver slows down before turning, but only that deceleration occurred. Once learned, these motifs can be used to improve one's estimates of the states themselves. This context-aware motif-based assignment is especially useful for adjusting to noise in the data. For example, if a noisy measurement is (incorrectly) assigned to the wrong state, yet the sequence it is in is ``close to'' being an instance of a known motif, one can use this motif to re-assign the measurement to the correct state (Figure \ref{Fig1}). Identifying an informative set of motifs is challenging because we cannot simply rank motifs by frequency: one seeks a set of motifs that are strong but are not redundant. Even ranking motifs is difficult, because one must consider both the significance of the motif pattern and the strength of the motif's instances in the dataset. %In this paper, we describe novel approaches both for curating a set of motifs and ranking motifs against each other.

Discovering motifs in temporal datasets requires an unsupervised way of locating and labeling repeated behaviors. In general, as the states themselves are unknown, methods must simultaneously uncover the states, assign states to measurements, and identify the repeated motifs. Unlike standard time series segmentation techniques~\cite{ggs,himberg2001time}, motifs repeat themselves many times across the time series. In contrast to clustering-only methods for state detection, many segments can be assigned to the same state~\cite{relatedHMM,dtw}. Additionally, existing model-based approaches for time-series motif discovery treat motifs as a final step and do not allow such motifs to influence and improve the robustness of the state assignment~\cite{sax}.

Here we introduce \emph{Motif-Aware State Assignment} (MASA), a method for discovering noisy motifs in time series data which are used to robustly assign states to the underlying measurements. We optimize over three types of parameters: the first defines a model over $K$ unique states, the second assigns each measurement to one of these states, and the third learns a set of motifs over the state assignment. Since these variables are combinatorial and coupled together in a highly non-convex manner, we solve the MASA optimization problem using an expectation-maximization (EM) type algorithm. Fixing our state model, we discover motifs and leverage them to robustly assign measurements to states. Then, fixing the state assignment, we update our state model. By iterating between these two steps, MASA uses motifs to shift the state models to include noisy measurements in order promote the repetition of global trends. During this process, MASA leverages novel techniques to scalably discover and curate a set of candidate motifs that are significant while minimizing redundancy.

We evaluate MASA on several synthetic and real-world datasets. First, we analyze how accurately MASA is able to label a time series with known ground-truth states. We compare our method with several state-of-the-art baselines, showing that MASA outperforms the best of the baselines by up to 38.2\%. We further assess MASA's ability to identify planted motifs in the midst of irrelevant datapoints, and find that MASA can accurately isolate these motifs. We also examine how robust MASA is to the presence of noisy data, measuring how performance fluctuates with varying amounts of noise. Overall, we find that MASA outperforms existing methods in terms of accuracy in state assignment and robustness to noise in the dataset. We validate these results by repeating the experiment with real sensor data reported from subjects cycling on an exercise bike. Finally, we perform two real-world case studies by applying MASA to sensor data from both aircrafts and automobiles. We show that MASA discovers interesting and interpretable motifs that occur in flight and driving sessions.

\xhdr{Further Related Work}
Recent approaches to time series clustering and segmentation often use distance-based metrics to identify different states~\cite{dtw}. Additionally, they apply dimensionality reduction~\cite{sax,symbolicMethod} or rule based approaches~\cite{ruleDiscovery,visualizing} to identify symbolic representations of time series data.
However, distance-based methods have been shown to be unreliable in certain cases~\cite{distanceMeaningless}, and lose the interpretability of multivariate data-points. Model-based methods, such as TICC~\cite{TICC}, ARMA~\cite{ARMA}, Gaussian mixture models~\cite{gaussianMixture}, and hidden Markov models~\cite{relatedHMM}, represent states as clusters using probabilistic models, and often can more accurately encode the relationships between sensors and the true underlying states. However, these existing methods have no way of incorporating motifs into their models.

Motif discovery is a common problem in time series data analysis~\cite{probablisticMotifs}. Methods for finding motifs include random projection~\cite{randomProjection} and suffix arrays~\cite{ACME}. Some of these methods are event rather than numerically based and thus bypass the simultaneous problem of state assignment ~\cite{papapetrou2005discovering}. Most of these methods assume motifs of fixed length ~\cite{grabocka2016latent}. Some methods also use distance metrics~\cite{uniformScaling}. The problem of finding repeated patterns also appears in the field of computational genomics. ACME uses a combinatorial approach to find super-maximal motifs in DNA~\cite{ACME}. Other bioinformatics models use edit-distance approaches to find motifs that vary slightly in appearance over the course of the sequence~\cite{editDistance}.
MASA departs from these methods in two respects. Firstly, MASA allows subsequences within a motif to have variable length while maintaining a given state. Moreover, unlike uniform scaling approaches \cite{uniformScaling}, MASA allows each state within a motif to scale independently. Secondly, MASA iterates by using the motifs to re-assign the original measurement to the updated states, encouraging noisy sequences to match a given motif. This iteration allows for stronger motifs and a more robust state definition.

\section{Problem Overview}
\label{setup}
Our input is a sequence of $T$ measurements, $\mathbf{X} = X_1,...,X_T$, where each measurement $X_t \in \mathbb{R}^N$ is a vector of data values observed at time $t$.
Our goal is to discover frequently occurring high level patterns, called motifs, in the input data.
A key component of our approach are $K$ state models which represent $K$ different states, where each state captures the properties of similar measurements.
We utilize these models to assign a state to each measurement and use the resulting state assignment to discover motifs.

%Our method works as follows.
%During the initialization, we compute the state models.
%We then utilize an expectation-maximization (EM) type approach to discover the motifs.
%In the expectation step (E-step) of EM, we first assign a state to each measurement, based on the state models.
%Next, we discover candidate motifs and then use discovered motifs to reassign state assignments.
%In the maximization step (M-step) of EM, we adjust state models to best match the reassigned measurement states.
%We iterate over EM steps until the state assignment converges at which point the candidate motifs contain the final output motifs.
%% Our goal is to develop models $\Theta_1,...,\Theta_K$ for $K$ states, and assign each measurement $X_i$ to a state. We can thus reduce our multivariate temporal dataset $\mathbf{X}$ into a symbolic string $\mathbf{S}$, where $S_i$, the $i$'th letter of the string, corresponds to the state assigned to $X_i$. Our algorithm will follow a typical EM flow. In the ``E''-step, we will assign each measurement to a state. In the M step, we will adjust our models $\Theta_1,...,\Theta_K$ to best explain the measurements that have been assigned to each state.
%MASA thus iteratively refines the state models, state assignments, and motifs, which allows for a robust motif discovery in the presence of noisy data. %Next, we describe the details of our approach.

\xhdr{Motifs}
\label{motifsection}
Our aim is to discover motifs that identify significant recurring and length-varying patterns in time series data.
We assume that these patterns contain time consecutive measurements.
Given a sequence of consecutive measurements, we define a $\emph{motif}$ as a sequence of corresponding state assignments, where all neighboring occurrences of the same state are merged into one. To illustrate, a car turn can be viewed as three states: (A) slowing down, (B) turning, and (C) speeding up. Although 
in a given instance of the turn, each of these states might have a different duration, the ``turn'' motif is represented by  $[A,B,C]$. 
Thus, states in the motif are ordered but the number of consecutive occurrences ({\emph{i.e.}}, duration) of each state may vary between instances of the motif.

%Symbolic representations of temporal datasets allow us to represent complex data relationships by a single label. However, the order in which states appear in the dataset may be even more indicative than the state assignment alone. We thus define a $\emph{motif}$ as a list of states describe a commonly occurring sequence in $\mathbf{S}$. The motif $[A,B,C]$ maps to a consecutive sequence of $A'$s, followed by $B$'s, followed by $C'$s. Note that motif does not specify the number of measurements in each consecutive sequence. Intuitively, if $[A,B,C]$ represented a driving turn sequence of slowing down, turning, and speeding up, the duration of time spent slowing down does not impact the classification of that slice of time as a turn.

%Motifs can thus be viewed analogously to regular expressions on the string $\mathbf{S}$. The motif $[A, B, C]$ corresponds the regular expression $A^*B^*C^*$: matches of this RegEx in $\mathbf{S}$ correspond to instances of the motif. Each motif can be thus represented by the pair $(m,q)$, where $m$ is the motif and $q$ is an associated list of its instances. A motif instance indicates one occurrence of the motif in the dataset.

Since we are interested in commonly occurring motifs, we represent each motif by the pair $(m,q)$, where $m$ is the motif and $q$ is an associated list of motif instances. A motif instance indicates one occurrence of the motif in the dataset.

We introduce the following motif constraints in our method:
\begin{enumerate}
\item A motif $m$ must contain at least 3 states: $|m| > 2$.
\item A motif $m$ must appear at least $L$ times: $|q| \geq L$.
%(for the rest of this paper, we set $L$ to a fixed value of $10$).
\item Motif instances cannot overlap: each measurement can only belong to at most one motif.
\end{enumerate}

We motivate the first constraint as motifs with two or fewer states are not very informative beyond the clustering itself. The second constraint aids our runtime: we do not want to spend time on investigating motifs that are not frequent. Since we are only interested in frequent patterns, we do not require that every measurement belongs to a motif:

%Importantly, not every measurement in our dataset needs to be associated with a motif: there may be some measurements $S_i \in \mathbf{S}$ which do not appear as part of any motifs. We slightly abuse notation by letting $\mathbbm{1}S_i \in \mathbf{M}$ be an indicator variable evaluating if $S_i$ is assigned to any motif.

%Motifs serve two purposes. They formalize a hierarchical structure on top of our symbolic representation, which allows us to identify significant recurring and length-varying patterns in time series data.
%However, we also leverage motifs in the $E$ step of our algorithm to provide better assignments of measurements to states. Intuitively, an assignment that allows a sequence of measurements to fit a known motif should be more likely. If a few noisy measurements would disrupt an otherwise strong instance of a motif, we should alter the assignment of those noisy measurements. At the same time, we do not want to force a measurement to be assigned to a state if, according to its data values, such an assignment is unlikely. We formalize this trade-off via the objective function below.

\xhdr{MASA Problem Setup}
Overall, MASA seeks to solve for a state model $\Theta$,
an assignment of states to measurements $\mathbf{S}$, and an assignment of motifs to measurements $\mathbf{M}$ to optimize the following objective (subject to motif constraints in the previous section):
$$\underset{\Theta, \mathbf{S}, \mathbf{M}}{\text{max.}} \sum_{i=1}^T \bigg( \log P_\Theta(X_i \mid S_i)
 - \beta \mathbbm{1}\{S_{i-1} \neq S_i\}
 + \log \gamma \mathbbm{1}\{S_i \notin \mathbf{M} \}\bigg)$$
$$
+ \Psi(\mathbf{M}) - R(\Theta).
$$
Here, $X_i$ is the measurement at time $i$, $S_i$ is its assigned state, and the probability $P_\Theta(X_i \mid S_i)$ is defined by our state model. The $\beta$ term is a hyperparameter that encourages neighboring measurements to be assigned the same state. The $\gamma$ parameter, $0 \leq \gamma \leq 1$, defines the cost of not assigning a measurement to a motif instance. Lower values of $\gamma$ indicate a more aggressive penalty for a measurement not conforming to any motif.
The term $\Psi$ is a scoring metric quantifying the strength of our motifs based on their occurrences in the dataset. %We give further discussion of this metric in Section \ref{assignment}.
 $R(\Theta)$ is a regularization penalty on our state model parameters $\Theta$. %; we discuss our state and state model in more detail below.

\xhdr{State model}
MASA is agnostic to the specific parameterization $\Theta$ of the state model $P_\Theta(X_i\mid S_i)$, which models the likelihood of an observation given a state.

MASA requires the following operations:
\begin{itemize}
    \item $P_\Theta(X_i\mid S_i)$: Distribution of how likely the measurement $X_i$ is observed given that it belongs to the state $S_i$.
    \item UpdateStatesModel($\mathbf{S}$): Once the states are assigned measurements, the state model parameterized by $\Theta$ must be updated to maximize the likelihood of the measurements conditioned on the state assignment $\mathbf{S}$. UpdateStatesModel optimizes this objective:
    $\max_{\Theta} \sum_{i=1}^T \log P_\Theta(X_i\mid S_i) - R(\Theta).$
    \item ProposeAssignment($\Theta$): Returns the state assignment $\mathbf{S}$  optimizing the non-motif objective:
    $\max_{\mathbf{S_i}} \sum_{i=1}^T  \log P_\Theta(X_i\mid S_i) - \beta \mathbbm{1}\{S_{i-1} \neq S_i\}.$
\end{itemize}

Given a state likelihood model $P_\Theta$, we note that ProposeAssignment can typically be implemented via the Viterbi algorithm as in~\cite{jurafsky}. Thus, any potential state model can be used, which allows MASA to be applied to diverse types of data, from heterogeneous exponential families to categorical distributions~\cite{park2017learning}.

In this paper specifically, we define our state model using the Toeplitz Inverse Covariance-based Clustering (TICC) model \cite{TICC}. TICC defines each state by a block Toeplitz Gaussian inverse covariance matrix $\Theta_k \in \mathbb{R}^{N \times N}$ with empirical mean $\mu_k \in \mathbb{R}^N$. The probability $\log P(X_i\mid S_i)$ is then
\begin{equation}\label{eq: LL_TICC}
    \log P(X_i\mid S_i) = -(X_i - \mu_{S_i})^T \Theta_{S_i} (X_i - \mu_{S_i}) + \log \det \Theta_{S_i},
\end{equation}
and $R(\Theta)$ is an $\ell_1$ penalty on the inverse covariance matrices $\Theta_k$. UpdateStateModel is then simply a single iteration of TICC, and ProposeAssignment uses the Viterbi algorithm to find the most likely sequence of states.

\xhdr{Hyperparameters}
Overall, our MASA algorithm contains four hyperparameters: $K$, the number of states, $L$, the minimum number of instances for a motif to be considered valid, $\beta$, the switching penalty, and $\gamma$, the aggressiveness of how strongly we encourage measurements to conform to a motif. All MASA hyperparameters can be pre-defined by the user, tuned by hand, or chosen more systematically using a method such as BIC. For our experiments, we chose $L$ to be a set, consistent number ($L=10$) which gives us a reasonable runtime, and picked $K$ and $\beta$ via BIC; we discuss hyperparameter tuning for our experiments in more detail in Appendix B. We discuss how $\gamma$ can be chosen in Section \ref{syntheticdatasection}. The values of each of the hyperparameters for each of our experiments can be found in the appendix. 

\section{MASA Algorithm}
\label{algo}

%Our method works as follows.
%During the initialization, we compute the state models.
%We then utilize an expectation-maximization (EM) type approach to discover the motifs.
%In the expectation step (E-step) of EM, we first assign a state to each measurement, based on the state models.
%Next, we discover candidate motifs and then use discovered motifs to reassign state assignments.
%In the maximization step (M-step) of EM, we adjust state models to best match the reassigned measurement states.
%We iterate over EM steps until the state assignment converges at which point the candidate motifs contain the final output motifs.
%MASA thus iteratively refines the state models, state assignments, and motifs, which allows for a robust motif discovery in the presence of noisy data. 

% \begin{algorithm}[t!]
%     \textbf{Step 0:} Initialize $\mathbf{S}$\;
%     \Repeat{Converged}{
%         \textbf{Step 1:} Generate motif candidates based on $\mathbf{S}$ (Section \ref{candidates})\;
%         \textbf{Step 2:} Reassign measurements to states, recomputing $\mathbf{S}$ and $\mathbf{M}$ (Section \ref{assignment})\;
%         \textbf{Step 3:} Recompute state definitions
%     }
%     \Return{$\mathbf{S}$, $\mathbf{M}$}
%     \caption{MASA Overview}
%     \label{overviewalg}
%  \end{algorithm}

Our problem is non-convex, and solving for a global optimum is intractable. However, we solve this objective via an iterative EM-like alternating maximization approach. In the E-step, we assign states to measurements and discover motifs which we then leverage to robustly update state assignment. In the M-step, we then use the updated state assignment to recalculate the state probability model by updating $\Theta$. Upon convergence, state assignments $\mathbf{S}$ and identified motifs $\mathbf{M}$ are provided as method results. MASA thus iteratively refines the state models, state assignments, and motifs, which allows for a robust motif discovery in the presence of noise.

The broad outline of our method is described below, and we give further details in the following sections:

\xhdr{Initialize $\Theta$} Initialize the state model $\mathbf{\Theta}$ in a reasonable manner. We initialize by alternately calling UpdateStateModel($\mathbf{S}$) and ProposeAssignment($\Theta$) until convergence.

\xhdr{E-Step: Compute $\mathbf{S}$ and $\mathbf{M}$}
Use the state model to compute a motif-aware state assignment. This step proceeds in two phases.

\textit{E-Step A: Discover candidate motifs.} 
Use ProposeAssignment($\Theta$) to compute a preliminary state assignment $\mathbf{S}$, then discover repeated patterns in $\mathbf{S}$ and generate a set of candidate motifs (Section \ref{candidates}).

\textit{E-Step B: Reassign states to measurements using motifs.}
Using the candidate motifs, reassign the states to measurements in order to match discovered motifs. If a measurement does not belong to a motif, a user-defined cost $\gamma$ is imposed. We use a hidden Markov model (HMM) to model each motif separately, finding sequences of measurements that may conform to each motif. Since each measurement can only belong to a single motif, we aggregate the HMM output to obtain a state assignment $\mathbf{S}$ while maintaining the constraints described in Section \ref{setup}. This step gives us a new state assignment $\mathbf{S}$ as well as a set of motifs $\mathbf{M}$
(Section \ref{assignment}).

\xhdr{M-Step: Use $\mathbf{S}$ to recompute $\Theta$}
As the state assignment has changed, use UpdateStatesModel($\mathbf{S}$) to update the state model $\Theta$ and then jump back to the E-Step.

We can either repeat the E and M steps of MASA for a set number of iterations, or stop when MASA has converged. MASA converges if the state assignment $\mathbf{S}$ returned from E Step is the same as the state assignment returned by ProposeAssignment($\Theta$) on the updated $\Theta$ from the M Step. In that case, the optimal state assignment $\mathbf{S}$ according to the state model is the same as the one suggested by the motifs, which means that our model reflects our motif preferences.

Next we give more details about each of the step of our algorithm.

\section{E-Step A: Discover candidate motifs}
\label{candidates}

Given $\Theta$, we compute a preliminary assignment of states to measurements by setting $\mathbf{S} = \text{ProposeAssignment}(\Theta)$. Using letters to indicate states, we can view $\mathbf{S}$ as a string, where each letter denotes a state assigned to a measurement. 
Because our motifs are independent of the number of adjacent equal states, our method operates on $\mathbf{S'}$, which is obtained from  $\mathbf{S}$ by collapsing consecutive duplicate letters into a single letter.  For example, a state assignment $\mathbf{S} =$ [A,A,B,B,B,B,C,C,C,C] gets collapsed to $\mathbf{S'} = $ [A,B,C].
%We thus seek to find common patterns of the form $A^*B^*\dots$.

%Because motifs do not specify the length of each consecutive subsequence, we operate over $\mathbf{S'}$ where consecutive duplicate letters in $\mathbf{S}$ have been collapsed.

\xhdr{Identify Repeated State Sequences}
We first seek to find all maximal repeated subsequences of states in $\bbS'$. A maximal subsequence is defined as a sequence that cannot be extended to either the left or right without changing the set of occurrences in $\bbS'$~\cite{ACME}. We require that each repeated  subsequence has at least $L$ non-overlapping instances in $\bbS'$. We use a suffix array to solve this problem efficiently~\cite{pyrstrmax}.  

\xhdr{Select Candidate Motifs} 
Each repeated subsequence of states in $\bbS'$ is a motif candidate. We need to select the relevant candidates for the final candidate set. We cannot simply select the motifs by frequency because an ideal motif candidate should also be a novel addition to the set. The motif $[A, B, C, D]$  may be largely useless if we already accepted $[A, B, C]$. We propose a dynamic way to select motif candidates by assessing each new candidate against a null model with knowledge of all previously accepted candidate motifs.

\xhdr{Motif occurrence probability} According to the null model, let the probability of observing an instance of the motif $m$ be the probability of independently observing each of $m$'s states $m_i$ according to their empirical frequency in $\bbS'$. If there are $N_m$ instances of the candidate motif in $\bbS'$, then the $p$-value of $m$ is the probability of at least $N_m$ occurrences of $m$ appearing according to our null model under a binomial distribution \cite{pValueComputation}. We then select motifs that exceed a required threshold $\alpha$. However, this method alone does not protect against redundancy: we want to consider a candidate motif relative to the other candidates we have accepted. To do so, we define a set $\mathcal{D}$ as the set of candidate motifs that the null model ``already knows'' about. 

\xhdr{Dynamic candidate selection} We assess each candidate motif $m$ in order of increasing length. If any candidates $d \in \mathcal{D}$ appear as a substring of $m$, we replace that substring in $m$ with a single state that appears (according to the null model) with $d$'s empirical probability in $\bbS'$. We can then evaluate the probability according to the null model above. Intuitively, the null model will perform better given knowledge of the candidate sets versus a null model which still treated each state as independent. We elaborate on implementation details of this algorithm in Appendix \ref{candidateAppendix}.

Let $P_\emptyset(m \mid \mathcal{D})$ be the probability of motif $m$ according to the null model with knowledge of $\mathcal{D}$. If for candidate motif $m$
\begin{equation*}
    P\bigg(
        \mathcal{B}(|\bbS'|, P_\emptyset(m \mid \mathcal{D}) \geq N_m \bigg) \leq \alpha,
\end{equation*}
for threshold $\alpha$, we accept $m$ as part of our candidate set, and add $m$ to $\mathcal{D}$ for evaluation of the next candidate. After this step, we have collected a novel and relevant set of candidate motifs, which we leverage in the next step to robustly assign states to measurements.

\section{E-Step B: Using Motifs to Assign States}
\label{assignment}

Having collected a set of candidate motifs, we next seek to assign a state to each measurement in order to promote these candidate motifs. Simultaneously, we decide on a final set of motifs $\mathbf{M}$ that fit the constraints laid out in Section~\ref{setup}. This step proceeds in 4 phases.

\xhdr{Identify New Motif Instances}
In this step, we identify possible instances of each candidate motif in our measurements. We specifically are looking for \emph{noisy} instances of the motifs which could not be found from the symbolic output of the non-motif optimization ProposeAssignment($\Theta)$. For example, if under our current state model $X_i$ is most likely in state $j$, but assignment to state $k$ would allow for completion of a motif with only a minimal effect on our likelihood objective, we should assign $X_i$ to $k$ instead of $j$.

Jointly considering all of our candidate motifs from E-Step A when re-assigning our measurements is both memory-expensive and slow. Instead, we consider a single candidate motif at a time, and re-assign all measurements according to this one motif to find a complete set of possible instances for this motif in the dataset. This approach simplifies our model and allows us to parallelize re-assignment by modeling every candidate motif concurrently.

\begin{figure}[!th]
    \centering
    \begin{tikzpicture}[->,>=stealth',shorten >=1pt,auto,node distance=2.2cm, semithick]
    \tikzstyle{every node}=[font=\footnotesize]

      \node[accepting, state]      (G)              {$z_0: \emptyset$};
      \node[state]     (A) [right of=G] {$z_1: A$};
      \node[state]              (B) [right of =A] {$z_2: B$ };
      \node[accepting, state]   (C) [right of =B] {$z_3: C$};

      \path (G) edge [loop above, looseness=5] node {$\beta_{\emptyset}(t)$} (G)
                edge              node {$\beta$} (A)
            (A) edge [loop above, looseness=5] node {0} (A)
                edge              node {$\beta$} (B)
            (B) edge [loop above, looseness=5] node {0} (B)
                edge              node {$\beta$} (C)
            (C) edge [loop above, looseness=5] node {0} (C)
                edge [bend left=40] node {$\beta$} (G)
                edge [bend left=45, auto=right] node {$\beta$} (A);
    \end{tikzpicture}
    \vspace{-3mm}
    \caption{HMM for a motif $[A, B, C]$. Edges represent transition costs; $z_1,z_2,z_3$ are motif states, while $z_0$ represents measurements that are not part of a motif.}
    \label{fig:markovmodel}
    \vspace{-4mm}
    \end{figure}
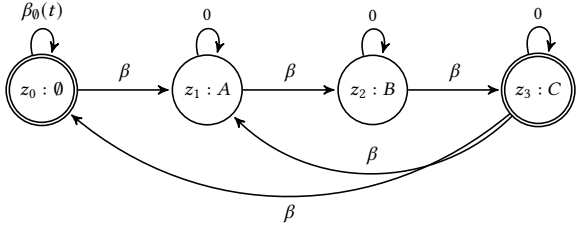

For each candidate motif $m$, we define a time-varying hidden Markov model (HMM) to model the entire sequence of measurements $\mathbf{X}$. Figure \ref{fig:markovmodel} shows an example of our HMM for $m=[A, B, C]$. The HMM has a hidden state $z_i$ for each state $m_i$ in $m$. Our emission probabilities for these hidden states are then
$
    \log P(X_i \mid z_j) = \log P_\Theta(X_i \mid m_j).
$
We further create a ``non-motif'' hidden state $z_0$, signifying that the measurement should not be assigned to this motif. We formalize $z_0$ as the original state assignment $S_i$, discounted by the non-motif penalty $\gamma < 1$. The emission probability for measurement $X_i$ from the non-motif state is then
$ \log P(X_i\mid z_0) =  \log P_\Theta(X_i \mid S_i) + \log(\gamma).
$

MASA prioritizes assignments of measurements to motif states even if a different assignment is more likely. $\gamma$ encapsulates how much less likely the motif state for a measurement $X_i$ can be while still being picked over the ``most likely'' sequence according to $X_i$'s values. A $\gamma = 1$ would assign all measurements to $z_0$, since the non-motif state already represents an optimal sequence. A lower $\gamma$ such as $\gamma = 0.6$ will be more aggressive, exploring the motif hidden states even when the likelihood might decrease.

Motif hidden states can transition only to themselves or into the next motif hidden state: In Figure~\ref{fig:markovmodel}, $z_1$ can transition to $z_2$ with penalty $\beta$, the state switching penalty. However, $z_1$ cannot directly transition to $z_3$. The non-motif hidden state can switch into itself or start a motif instance by switching into $z_1$; however, once $z_1$ has been entered, the only way to return to the non-motif hidden state is by finishing the motif. The last hidden state ($z_3$) can either start another motif by switching to $z_1$ or enter the the non-motif $z_0$.

States do not incur a switching penalty when switching to themselves. The notable exception is the non-motif state, where we define the penalty of looping in the non-motif state using $\beta_{\emptyset}(t)$:
$$
\beta_{\emptyset}(t) = \beta \mathbbm{1}(t\neq 1 \land S_{X_{t-1}} \neq S_{X_t})
$$
$\beta_{\emptyset}(t)$ thus adds a cost of $\beta$ when transitioning from $z_0$ to itself if, in our original $\mathbf{S}$, this measurement also incurred a switching penalty. Both $z_0$ and the first motif hidden state (in our example in Figure~\ref{fig:markovmodel}, $z_1$) are valid starting states. $z_0$ and the last motif state ($z_3$) are the only valid end states.

With our HMM fully defined, we use the Viterbi algorithm, a dynamic programming method, to find the most likely sequence of hidden states for these measurements \cite{jurafsky}. Any $X_i$ which are in the motif states represent a possible motif instance for $m$. Thus, after performing this step, we have a list of possible instances for every candidate motif $m$.

\xhdr{Scoring Motif Instances}
Since we build a separate HMM for each motif in the previous phase, it is possible for a single measurement to appear in multiple motifs, violating the restrictions in Section \ref{setup}. Therefore, it is necessary to develop a way of ranking motif instances and assigning each measurement to only one motif. Each instance has two factors influencing its significance: the importance of the motif itself and the individual likelihood of the measurements.

\textit{Motif score.} Let $N_m$ be the number of instances of motif $m$ identified via the HMM. We again define our null model under the assumption that each state in the motif occurs independently based on the empirical frequency counts from $\bbS'$ from Section \ref{candidates}. If $\mathbb{E}_\emptyset[N_m]$ is the expected number of instances of $m$ according to the null model, then we define the motif score as the G-score \cite{GTEST}:
$$ \Upsilon(m) = 2 N_m \left(  \log N_m - \log \mathbb{E}_\emptyset[N_m] \right). $$

\textit{Instance score.}
We now formalize the second component of our scoring metric, which evaluates an individual instance of a motif candidate. Let $X_i,...,X_j$ be a consecutive sequence of measurements which were identified as a possible instance of the motif $m$ by our HMM, where $S_i$ is the state assignment of $X_i$ according to $\mathbf{S}$ and $S^{(HMM)}_i$ is $X_i$'s new assignment according to the motif model. Then we compute the instance score as the log ratio between the likelihoods of the two assignments \cite{LRT}: 
$$\Delta(m, (X_i,...,X_j)) = \sum_{k=i}^j 2 \log P(X_i \mid S^{(HMM)}_k) - \log P_\Theta(X_i \mid S_k).$$
Then the total score for a motif instance is:
$$
\Psi(m,(X_i,...,X_j)) = \Upsilon(m) + \Delta(m, (X_i,...,X_j)).
$$
And our total motif score for $\mathbf{M}$ is:
$$ \Psi(\mathbf{M}) = \sum_{(m, q) \in \mathbf{M}} \Upsilon(m) + \sum_{i=1}^{|q|} \Delta(m, q_i).$$

\begin{algorithm}[t!] 

    Sort instances by $\Psi$; initialize $X_1,...,X_T$ open\;
    \ForEach{motif instance $(m,q)$} {
        \If{any $X_i$ covered by $q$ is locked} {
            reject $q$ and continue\;
        }
        \eIf{$m$ is complete} {
            Lock all measurements covered by $q$\;
            Remove other bids on those measurements\;
        }{
            Place bid on all measurements covered by $q$\;
            \If{\# instances bid by $m \geq L$}{
                Mark $m$ as complete\;
                Lock all of $m$'s current bids\;
                Remove other bids on those measurements\;
            }
        }
    }
    Retrieve motif instances that are permanently locked\;
    Construct and retrieve $\mathbf{M}$\;
    \caption{Greedy Motif Assignment}
    \label{greedyalg}
\end{algorithm}

\xhdr{Update State Assignment}
Having curated a set of motif instances and developed a scoring methodology, we now update the state assignment $\mathbf{S}$. We do this by allocating each measurement $X_i$ to either a single motif or no motif. If a measurement is assigned to motif $m$, then it takes its state according to $m$'s HMM. If a measurement is not assigned to any motif, it keeps its old state from the old $\mathbf{S}$. While performing this process, we need to uphold the constraints in Section~\ref{setup}.

After sorting all of the found motif instances from all motifs according to $\Psi$, we greedily allocate measurements to motifs using a system of locks and bids (Algorithm \ref{greedyalg}). Each measurement can be either ``locked'' or ``open''.
Initially, all measurements are open. A measurement becomes locked when it is permanently assigned to a motif. We enforce a constraint that multiple motifs can place a bid on an open measurement, but a locked measurement can only belong to a single motif. Only motifs that have at least $L$ instances can lock measurements; we label such motifs as ``complete''.

After processing all motif instances in Algorithm \ref{greedyalg}, we only accept assignments of locked measurements to motifs. Since only complete motifs can lock, only motifs with at least $L$ instances will appear in the final set of motifs. This lock/bid scheme further ensures that no $X_i$ is assigned to multiple motif instances. Any measurement that does not belong to an accepted motif instance is set to its original assignment in $\mathbf{S}$.

% Firstly, we set all measurements as open. We then sort the motif instances by their score $\Psi$: each possible motif instance $q$ is a consecutive sequence of measurements that we think might conform to motif $m$. For each instance $q$ belonging to motif $m$, check if any of the $X_i \in q$ are locked, in which case they reject $q$. If no measurement in $q$ is locked, we say that the underlying motif $m$ has ``bid", or temporarily claimed, segment; we then move on to the next motif instance $q$.

% If at any time a motif $m$ has bid on $L$ instances, we declare $m$ as complete. We permanently lock down all measurements in the instances that $m$'s has bids on: these instances will appear in the final assignment. We remove bids for any other motif who bid on those measurements. Moving forward, any $q$ that appears for $m$, if it does not infringe on locked space, can immediately lock its sequence of measurements without bidding.

We thus have a completely new assignment of measurements to states $\mathbf{S}$. We further construct $\mathbf{M}$ as the set of complete motifs.

\section{M-Step: Recompute state model}
\label{mstep}
By creating a new $\mathbf{S}$, we have shifted the pool of measurements assigned to each state to include additional measurements based on motif assignment. We thus now use UpdateStatesModel($\bbS$) to recompute $\Theta$.

We then check for convergence. Using our recomputed likelihood model, if ProposeAssignment($\Theta$) returns the same state assignment $\bbS$ that was returned from the E-Step, then our states have stabilized and MASA has converged. Otherwise we pass our updated $\Theta$ to the E-Step from the M-Step and repeat.

\section{Convergence and Runtime of MASA}
MASA converges when recomputing our state model $\Theta$ does not change our state assignment. MASA can alternatively be capped at a set number of iterations. The number of iterations that MASA needs to converge is dataset dependent, but typically is on the order of tens of iterations. The runtime of each iteration of MASA is:

\xhdr{E-Step A: Motif Discovery} For a time series of length $T$ and minimum motif length $L$, we identify at most $T/L$ patterns from the suffix array. Let $C$ be the number of eventually accepted motif candidates. Since construction of the suffix array takes linear time, generation of motif candidates takes worst case time $O((T/L)C + T)$. However, $C \ll T$ since motif construction occurs from the collapsed form $\bbS'$ and is further constrained by the significance threshold $\alpha$. We can optionally cap $C$ by truncating the set according to p-score. This step thus takes linear time in $T$.

\xhdr{E-Step B: Motif assignment and scoring} Assuming any motif candidate has at most $r$ states defining the motif sequence,  each HMM takes $O(rT)$ time due to the chain structure of the HMM. We can identify instances for each motif candidate in parallel. Scoring the instances thus takes constant time per motif instance. In the worst case for Algorithm \ref{greedyalg}, each motif has instances that cover all $T$ measurements. In such a scenario, each motif candidate can only bid on a measurement once, so the algorithm can take worst case $O(CT)$. Practically, however, this is a loose upper bound since such a scenario would indicate an extremely redundant candidate set and would not pass through the filter in E-Step A. Thus, this algorithm takes runtime linear in $T$. 

\xhdr{M-Step: Recompute State Model}
Finally, since UpdateStateModel is also linear in $T$, M-Step also takes linear time in $T$. 

Thus, in total, each iteration of MASA takes {\em linear time} in $T$, which we further verify experimentally in Section \ref{scalability}.

\section{Evaluation}
We performed an extensive evaluation of MASA's performance on a range of scenarios and different parameter values. Specifically, we seek to measure MASA's ability to identify common motifs in the presence of noise, as well as its accuracy in assigning points to the correct underlying states. For evaluations in this section, we initially use synthetic datasets, since they provide a known ground truth which we require for comparing MASA with state-of-the-art baseline methods. We then validate these results on real cycling data with planted motifs. Later, in Section \ref{casestudies}, we apply MASA to two real-world case studies. 

To run the experiments, we built a MASA solver that implements the algorithm described in Section \ref{algo}. A link to our solver can be found in the footnote.\footnote{Our MASA solver and all code from the experiments in this section are available for download at \url{https://github.com/snap-stanford/masa.git}.}. Given a multivariate time series dataset our solver returns the ranked motifs, the motif assignments, and the state assignments for each measurement.

\begin{figure}[t!]
    \centering
    \includegraphics[width=0.49\textwidth]{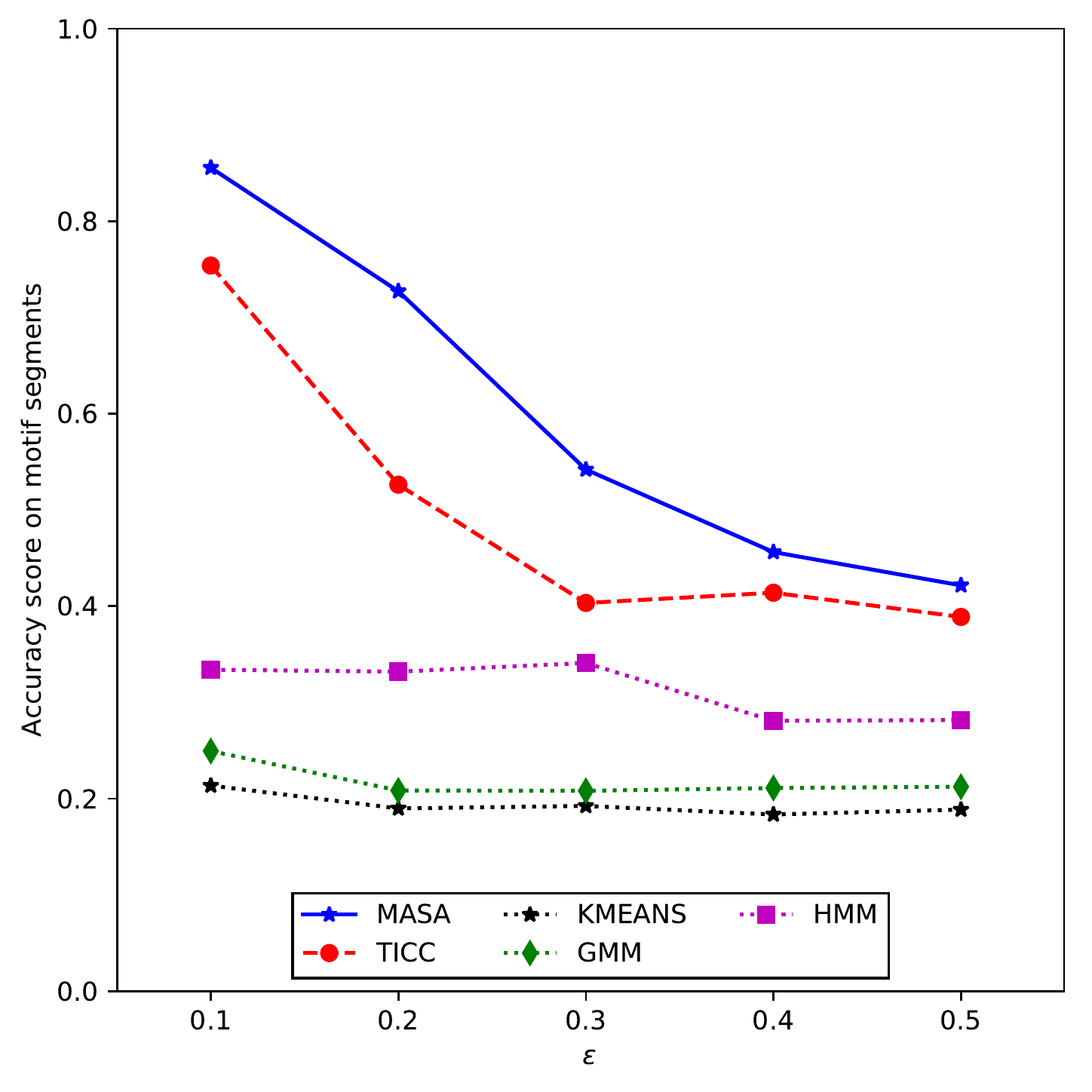}
    \vspace{-10mm}
    \caption{MASA and baseline accuracy scores on motif sections for synthetic data with varying levels of $\epsilon$.}
    \label{accuracy}
    \vspace{-4mm}
\end{figure}

\subsection{Experiments on Synthetic Data}
\label{syntheticdatasection}

\xhdr{Synthetic Dataset Generation} 
We generate our dataset as follows: Our ground-truth synthetic time series has a total of 150,000 measurements, each measurement being in $\mathbb{R}^5$. Measurements are taken from $K$ states; we use $K=10$ in our experiments. We first assign states to measurements and then generate specific data values for each measurement.

To create the assignment, we compose the time series from 1,000 \quotes{macro-segments}, each containing 150 measurements.
Each macro-segment begins with 6 \quotes{non-motif} segments of 15 measurements, where measurements within each segment are assigned randomly to one of $K$ states.
All measurements in one \quotes{non-motif} segment are assigned the same state. %, but measurements in different segments can be assigned to different states.
These non-motif segments are followed by 4 \quotes{motif} segments of 15 measurements each with states $A \rightarrow B \rightarrow C \rightarrow D$.  Using these ground truth assignments, we form random ground truth covariances $\Sigma_1,...,\Sigma_K$ for each state through the method described in \cite{mohan2014node}. Each data-point from ground truth state $k$ is then drawn from a multivariate normal distribution with zero mean and covariance $\Sigma_k$ as in \cite{TICC}.

We introduce noise into our segments as follows. For every segment, with probability $\epsilon$, we perturb measurements in that segment by a random non-motif state. For example, suppose a segment with a ground truth state $i$ is set to be perturbed. We pick a random state $j \notin \{A, B, C, D\}$ (the non-motif states). Rather than using $\Sigma_i$ to generate data, we draw from a distribution with covariance $0.7 \: \Sigma_{j} + 0.3 \: \Sigma_i$, such that the new segment
appears as if from the random state.

\xhdr{Robustness to Noise}
The synthetic dataset plants motifs with noisy segments. While other algorithms might struggle with these noisy segments, MASA can leverage the motif structure to smooth out the state assignment of these perturbed segments. 

To evaluate the effectiveness of MASA in noisy data, we create multiple datasets using the above method, each with a different $\epsilon$ value. A larger value of $\epsilon$ indicates greater noise in the data, since more measurement values will be perturbed. We run MASA at $\gamma = 0.8$ (later in this section, we evaluate the sensitivity of our results to the selection of $\gamma$) against TICC and the following baseline models: a Gaussian mixture model \cite{gaussianMixture}, K-means with Euclidean distance, and a hidden Markov model with Gaussian emissions.

Figure \ref{accuracy} shows the results from measuring the accuracy of these models on the motif sections of the dataset against the ground truth.  The accuracy quantifies the probability that
a given measurement was classified into the correct state.
%gives, for every measurement that was generated as part of a ground truth motif $A \rightarrow B \rightarrow C \rightarrow D$, the probability that this measurement was properly classified into the correct state.
As shown, MASA classifies these measurements with higher accuracy than any of the other baselines at all values of $\epsilon$, outperforming TICC (the second-best performing method) by up to 38.2\%.
Interestingly, as $\epsilon$ rises initially up to about $0.3$, MASA outperforms TICC by an increasing margin, which means that MASA is able to successfully utilize context information to correct state assignments.
%As expected, when $\epsilon$ rises even further, the support of $A \rightarrow B \rightarrow C \rightarrow D$ decreases, which limits MASA's ability to identify it as a strong motif and to correct assignments missed by TICC. The gap between MASA and TICC thus narrows with higher levels of $\epsilon$ as MASA corrects fewer state assignments in the presence of increasing amounts of noise.

\begin{table}
    \centering
    \caption{Weighted $F_1$ scores on motif states $A, B, C$ and $D$ as a function 
    of noise $\epsilon$. Notice that MASA performs best across all noise levels.}
\bgroup
\def\arraystretch{1}
\setlength\tabcolsep{3pt}
\begin{tabular}
{|r|| c c c c c|}
\hline
$\mathbf{\epsilon}$  & $\mathbf{0.1}$ & $\mathbf{0.2}$ & $\mathbf{0.3}$ & $\mathbf{0.4}$ & $\mathbf{0.5}$
\\ \hline  \hline
$\mathbf{MASA}$  & $\mathbf{0.759}$  &$\mathbf{0.719}$  &$\mathbf{0.518}$  & $\mathbf{0.464}$  &$\mathbf{0.440}$ \\ \hline
$\mathbf{TICC}$  &$ 0.736$  &$ 0.599$  &$ 0.419$  &$ 0.462$  &$ 0.436$ \\ \hline
$\mathbf{KMEANS}$  &$ 0.217$  &$ 0.208$  &$ 0.198$  &$ 0.192$  &$ 0.202$ \\ \hline
$\mathbf{GMM}$  &$ 0.234$  &$ 0.185$  &$ 0.190$  &$ 0.191$  &$ 0.219$ \\ \hline
$\mathbf{HMM}$  &$ 0.252$  &$ 0.269$  &$ 0.283$  &$ 0.256$  &$ 0.263$ \\ \hline
\end{tabular}
\egroup
% \vspace{2mm}
\vspace{-7mm}
\label{table:f1scores}
\end{table}

%To absorb planted noise, MASA must adapt the parameter values of states $A, B, C,$ and $D$, so that the noisy assignments are included. We seek to evaluate the impact of these adaptations over the entire dataset. 

We further compare methods on their ability to assign measurements to ground-truth motif states.
In Table~\ref{table:f1scores}, we evaluate the $F_1$ scores for states $A, B, C,$ and $D$ over the full dataset, including times where they are assigned in the non-motif sections. These scores are weighted by the support of these states within the ground truth dataset. MASA continues to significantly outperform TICC and other baselines, especially when the amount of noise is reasonable ($\epsilon < 0.4$). This shows how MASA can optimize accuracy on the motif sections while correctly ignoring the non-motif sections of the dataset.

%% MOTIF IDENTIFICATION
%\xhdr{Motif Identification}
%We further evaluate MASA's accuracy in identifying the entire sequence $A \rightarrow B \rightarrow C \rightarrow D$ as a motif. We run MASA for $\epsilon = 0.2$, labeling each measurement as 1 if it is included in the motif and 0 otherwise. The true labels thus assign each measurement as 0 unless it is part of the $A \rightarrow B \rightarrow C \rightarrow D$ sequence. The $F_1$ score then measures how accurately we assign measurements in the ground truth motif to the identified motif, while also ensuring that we do not incorrectly assign measurements in the ``non-motif'' sections to this motif by accident.
%
%For this experiment, MASA achieves an $F_1$ score of 0.79, showing that it successfully achieves the balance of properly labeling the true, but noisy, motifs, while not being overly aggressive by wrongly attaching incorrect sequences to this same motif. We note that the baseline methods are unable to perform this operation of recovering specific motifs, so there are no comparable results for their performance on this task.

% \begin{figure}[t!]
%     \centering
%     \includegraphics[width=0.5\textwidth]{scalability.pdf}
%     \caption{MASA per-iteration runtime for a synthetic dataset of varying lengths. Our solver scales linearly for $T$, the number of timesteps.}
%     \label{scalabilityplot}
% \end{figure}

\begin{figure}
    \centering
    \includegraphics[width=0.5\textwidth]{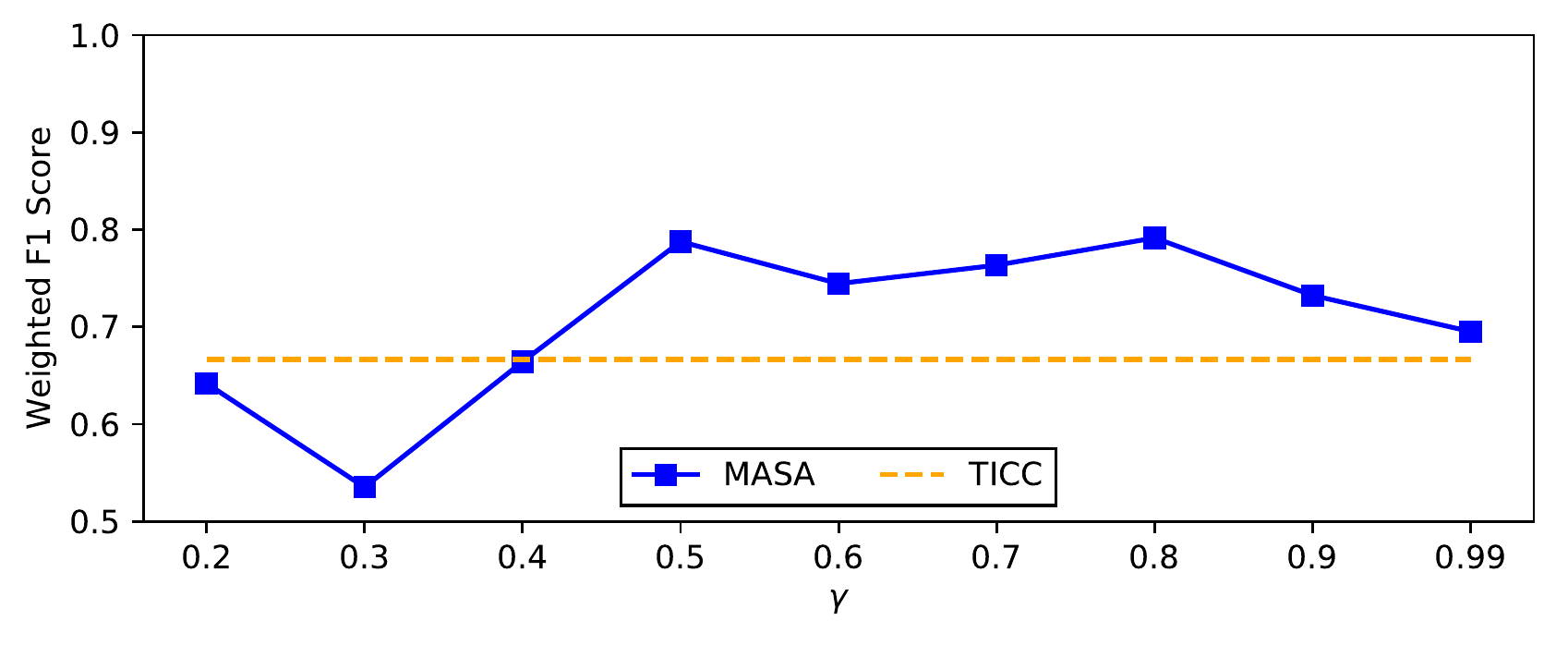}
    \includegraphics[width=0.5\textwidth]{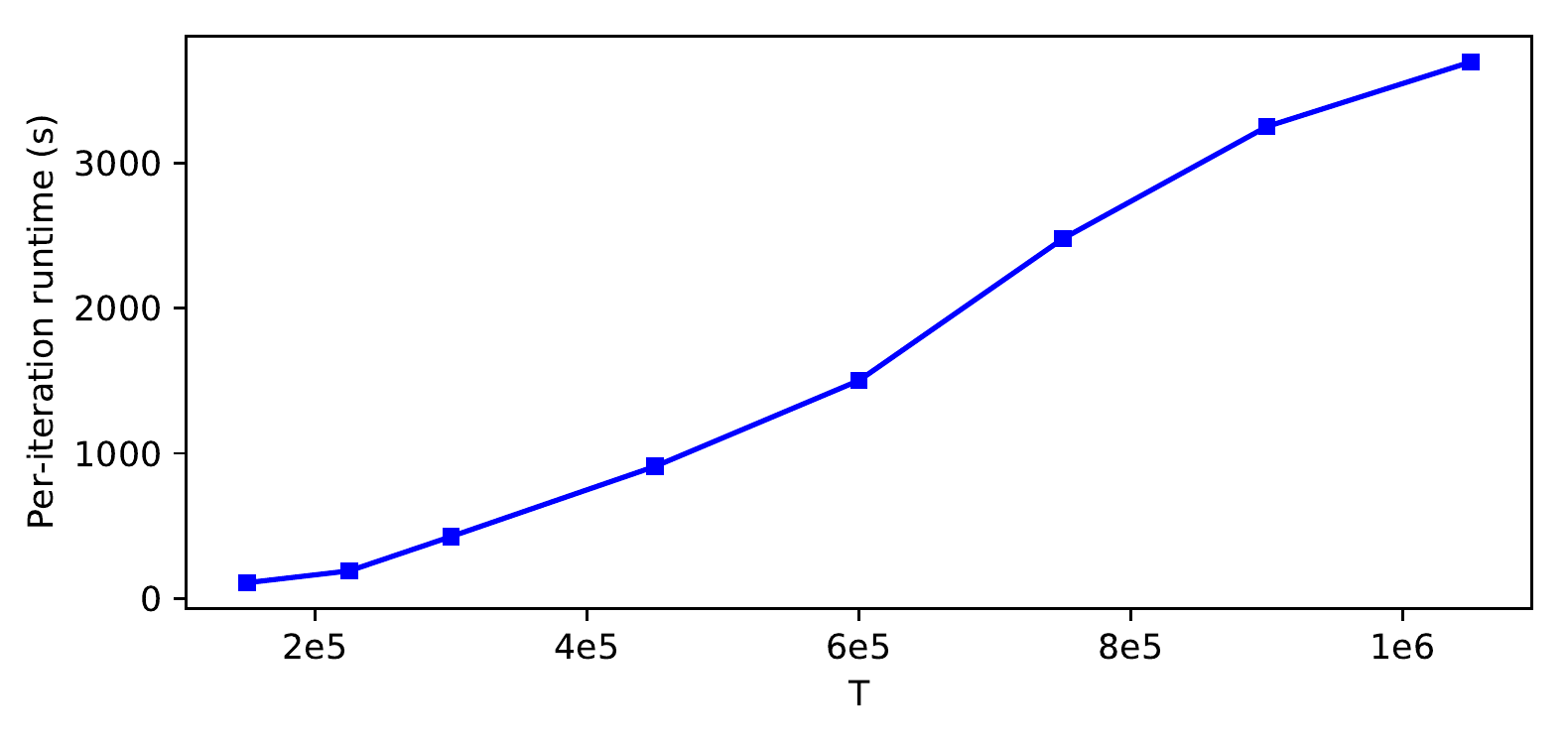}
    \vspace{-7mm}
    \caption{Top: Weighted $F_1$ scores on motif states for varying $\gamma$. Here, all values of $\gamma$ above 0.4 outperform TICC's $F_1$ score. Bottom: MASA per-iteration runtime for a synthetic dataset of varying lengths. Our solver scales linearly for $T$, the number of timesteps.}
    \vspace{-3mm}
    \label{robustness}
\end{figure}

\xhdr{$\gamma$-robustness}
We evaluate robustness to the parameter $\gamma$, which encapsulates the aggressiveness of ``forcing'' measurements to follow motifs. Smaller $\gamma$'s encourage sequences to conform to known motif patterns, even if they are not a perfect fit. Higher values of $\gamma$ only encourage motifs to exist when there is a near-perfect alignment. We run MASA with $\epsilon = 0.2$ and plot the weighted $F_1$ score on the motif states $A, B, C,$ and $D$ (including times where they are assigned to non-motif sections, similar to Table~\ref{table:f1scores}) in Figure \ref{robustness} (top). We see that MASA is robust to the selection of $\gamma$, obtaining a weighted $F_1$ score of above 0.69, the score that TICC achieves, for all $0.4 < \gamma \leq 0.99$. Because the initial motif filtering step only permits the algorithm to pursue plausible motifs, even low values of $\gamma$
(which are overly aggressive) perform reasonably well because there are very few ``incorrect'' motifs that the dataset can be wrongly assigned to.
 Overall, this shows how MASA can discover accurate trends even if the aggressiveness parameter $\gamma$ is not perfectly tuned to its optimal value.

\xhdr{Scalability}\label{scalability}
Since time series datasets can be extremely long, the number of measurements $T$ dominates the runtime of MASA. We empirically measure the per-iteration runtime of MASA in Figure \ref{robustness} (bottom) on synthetic data with $\epsilon=0.2$ for increasing numbers of measurements.
For consistency between measurements, we cap our number of motifs to $25$ as described in Section \ref{candidates}.
We find that the growth in runtime empirically increases linearly with respect to $T$, solving a dataset of $1.05$ million measurements in $3,700$ seconds.. Thus, MASA can scalably handle long time series datasets, and can solve an iteration for a dataset of over 1 million measurements in approximately one hour.

\begin{table}
    \centering
    \caption{State assignment accuracy scores on motif sections for cycling data.}
\bgroup
\def\arraystretch{1}
\setlength\tabcolsep{3pt}
\begin{tabular}
{|r|| c c c c c |}
\hline
 ~& $\mathbf{MASA}$  & $\mathbf{TICC}$ &$\mathbf{KMEANS}$  & $\mathbf{GMM}$ & $\mathbf{HMM}$ \\ \hline  \hline
Accuracy & $\mathbf{0.830}$  & $0.741$ & $0.614$ & $ 0.812$ & $0.710$ \\ \hline
\end{tabular}
\egroup
\vspace{2mm}

\vspace{-4mm}
\label{table:cycling}
\end{table}

\subsection{Experiments on Cycling Data}
To conclude our evaluation, we evaluate MASA's performance on planted motifs on real cycling data. We use the publicly available Daily and Sports Activities dataset~\cite{altun2010comparative}. In this dataset, eight subjects cycled on an exercise bike, with 45 sensor values being recorded at 25 Hz. The dataset was broken down for each cyclist into five-second runs, or 125 sample long segments, where each of the eight cyclists generated 60 samples, for a total of 480 five-second segments in the dataset. We seek to evaluate MASA's performance on identifying the cyclist given a sample of data.

We plant motifs in a manner similar to the synthetic dataset above. We randomly sample six segments (each of length 125), choosing with equal probability any of the 480 samples, as the ``non-motif'' section. For the motif, we use the motif [A, B, C, D], with each state representing the first four cyclists in the dataset. To generate the motif, we draw one segment from each of these four cyclists in order (randomly picking one of that subject's 60 samples), and concatenate the 125-sample segments to form the motif [A, B, C, D]. We perform this procedure 100 times for a total of 125,000 datapoints. We then run MASA and evaluate accuracy scores on the motif sections, as in Figure \ref{accuracy}. We find that MASA outperforms all other baselines (Table \ref{table:cycling}) with an accuracy score of 0.830 as compared with TICC's 0.741. We cover more specific details on this experiment in our supplementary section. 

\section{Case Studies}
\label{casestudies}
Here, we run MASA on two real-world examples, using automobile and aircraft sensor data to demonstrate how our approach can be applied to discover insightful motifs.

\begin{figure}[t!]
    \centering
    \includegraphics[width=0.45\textwidth]{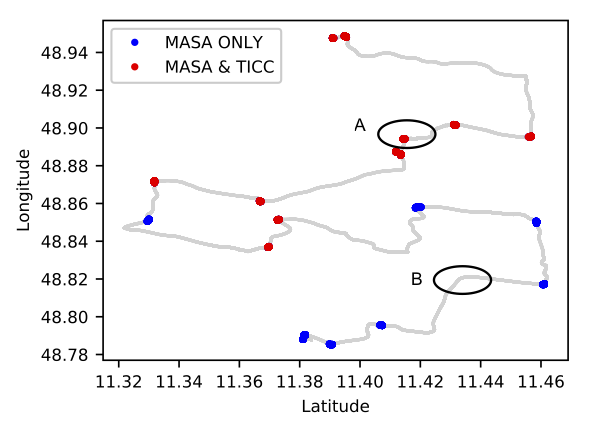}
    \includegraphics[width=0.5\textwidth]{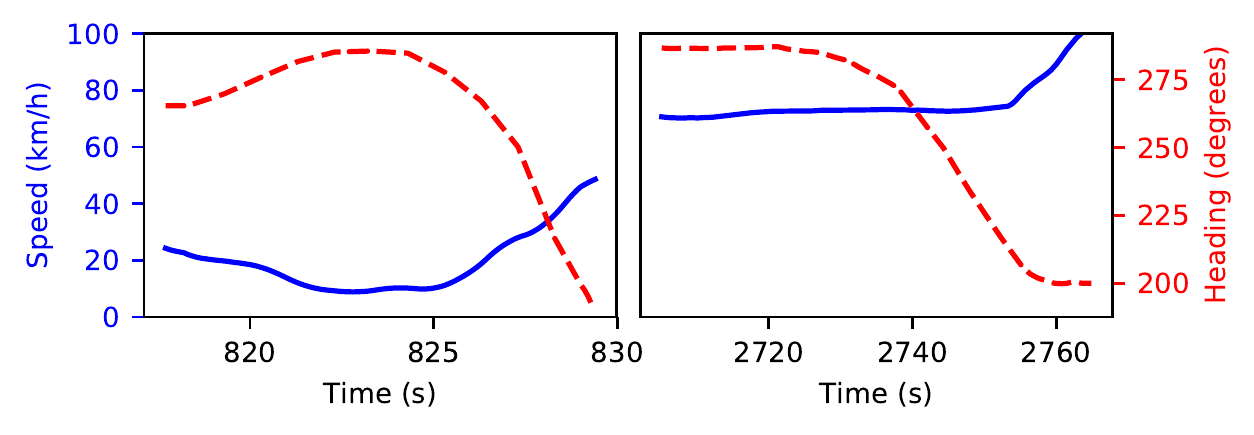}
    \vspace{-8mm}
    \caption{Top: Driver path over one session. Instances belonging to the \quotes{turn} motif are highlighted. Notice MASA identifies many more turns than TICC. Bottom: The speed/heading during Turn A (Left) and Turn B (Right).}
    \label{car}
\end{figure}

\xhdr{Automobile Sensor Data}
We first evaluate our algorithm on a multivariate car sensor dataset, provided by a large automobile company.  Data from 7 sensors was collected every 0.1 seconds over an hour of driving data on real roads in Germany: brake pedal position, acceleration (X-direction), acceleration (Y-direction), steering wheel angle, speed, engine RPM, and gas pedal position.

We run MASA with 8 states on this dataset, and the results immediately identify a significant three stage motif that occurs 19 times during the one-hour session. After plotting the GPS coordinates of each sensor reading in Figure \ref{car}, we see that this motif corresponds to the driver turning the vehicle. Specifically, the three states appear to correspond to ``slowing down'', ``turning the wheel'', and ``speeding up''. We then searched for this sequence of states in the TICC state assignment: we found that only 11 of the 19 turns (marked in Figure \ref{car}) identified by MASA conformed to the motif pattern. Moreover, the turn pattern did not otherwise appear in TICC's assignment. In these other cases, some noise in the sensor readings led TICC to assign one (or more) of the measurements in the turn to an incorrect state. However, using the turn motif, MASA is able to correct for such noise and identify the sequence as a turn in a completely unsupervised way.

There are some bends in the driving path which MASA does not identify as a turn. These are due to significant deviations in the ``typical turn'' identified by the motif. Figure \ref{car} depicts the speed and heading for the car during the section labeled A, which MASA identified as a turn. In this section, the car slows to 10 km/h during the turn before speeding back up. In contrast, during ``turn B", the car maintained a speed of at least 70 km/h during the entirety of the turn. This stretch of road occurred on a highway, so even though the heading of the vehicle changed, this section of road did not conform to a classic three-stage ``turn'' motif, and thus was not classified as a turn by MASA.

MASA can thus identify directly interpretable segments of interest in an automobile dataset. Compared to state-of-the-art methods like TICC, MASA can intervene to more robustly assign individual measurements to states to make common behaviors, such as turns, appear uniform throughout the dataset in the presence of noise.

\xhdr{Airplane Sensor Data}
\begin{figure}[t!]
    \centering
    \includegraphics[width=0.49\textwidth]{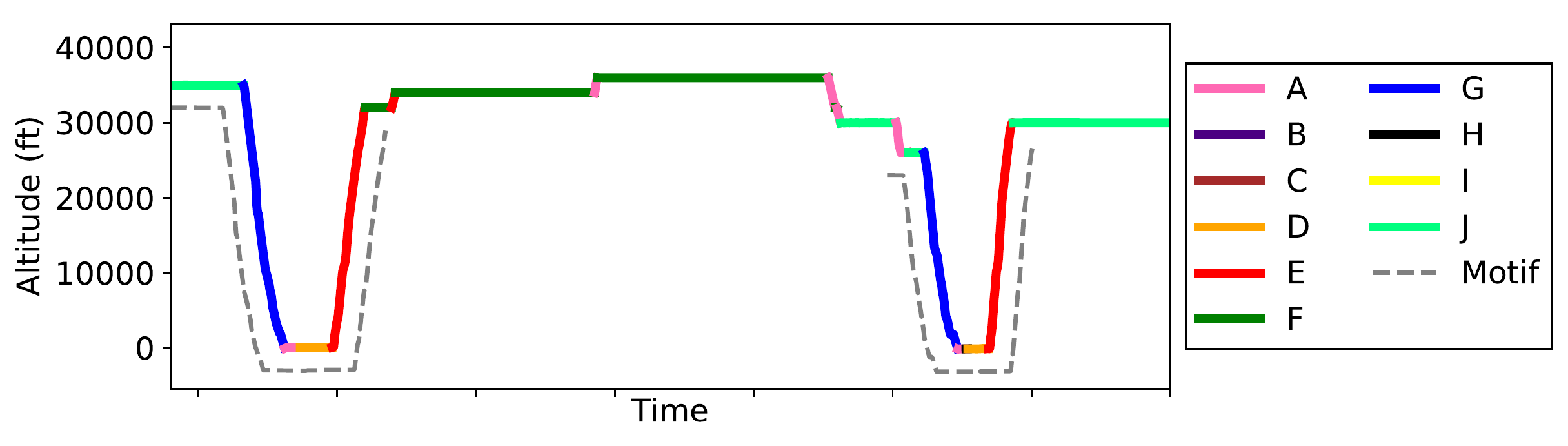}
    \vspace{-4mm}
    \caption{Airplane altitude over a 10 hour interval. The plot is colored according to its state assignment, and the specific motif instances are underlined.}
    \vspace{-4mm}
    \label{altitude}
\end{figure}
We next analyze a dataset, provided by a large manufacturer, containing data from a single commercial aircraft over the course of several months. This multivariate dataset contained 1,459 sensors collected over 85 total flights, where data was sampled every 10 seconds. For computational scalability, we embed each 1,459-dimensional measurement in a low-dimensional vector using principal component analysis. Here, we pick a value where the eigenvalues store 99\% of the cumulative energy, which yields a vector in $\mathbb{R}^{13}$. We then run MASA with 10 states.

\begin{figure}[t!]
    \centering
    \includegraphics[width=0.49\textwidth]{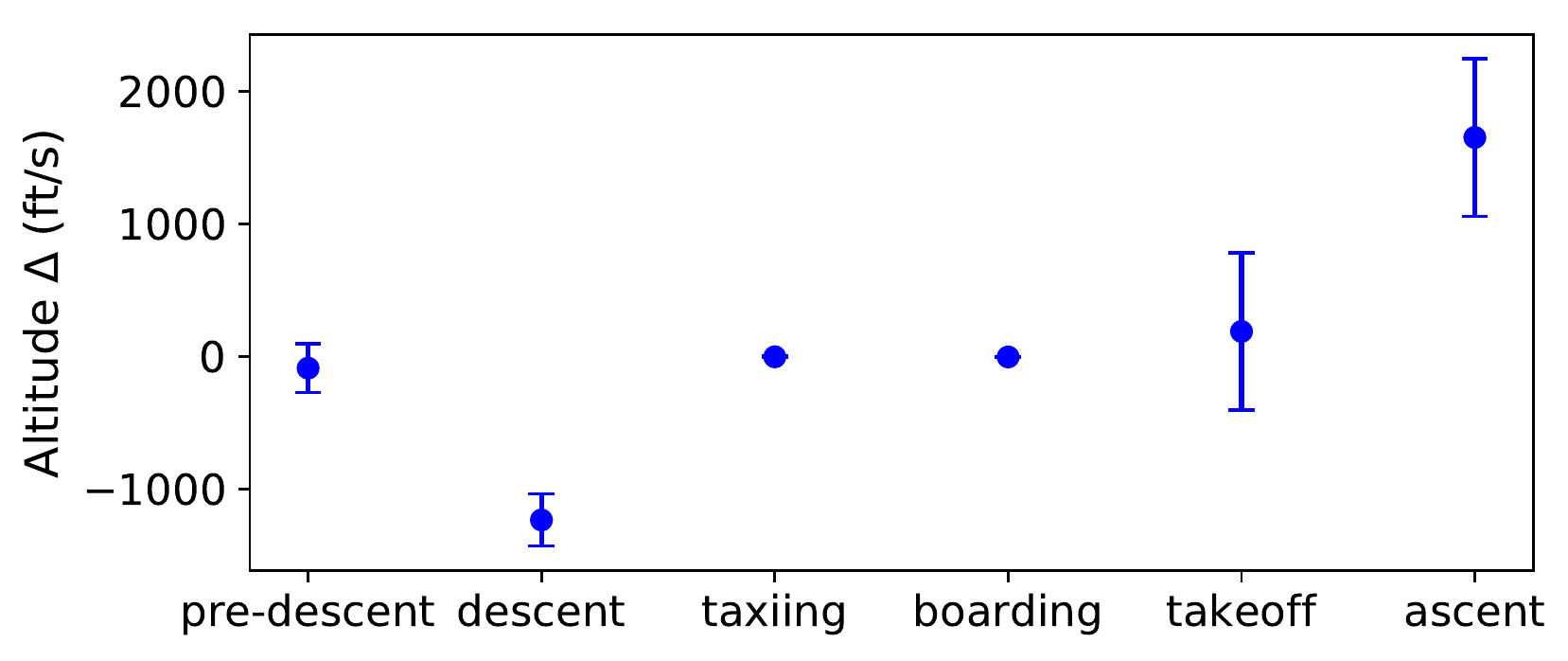}
    \includegraphics[width=0.49\textwidth]{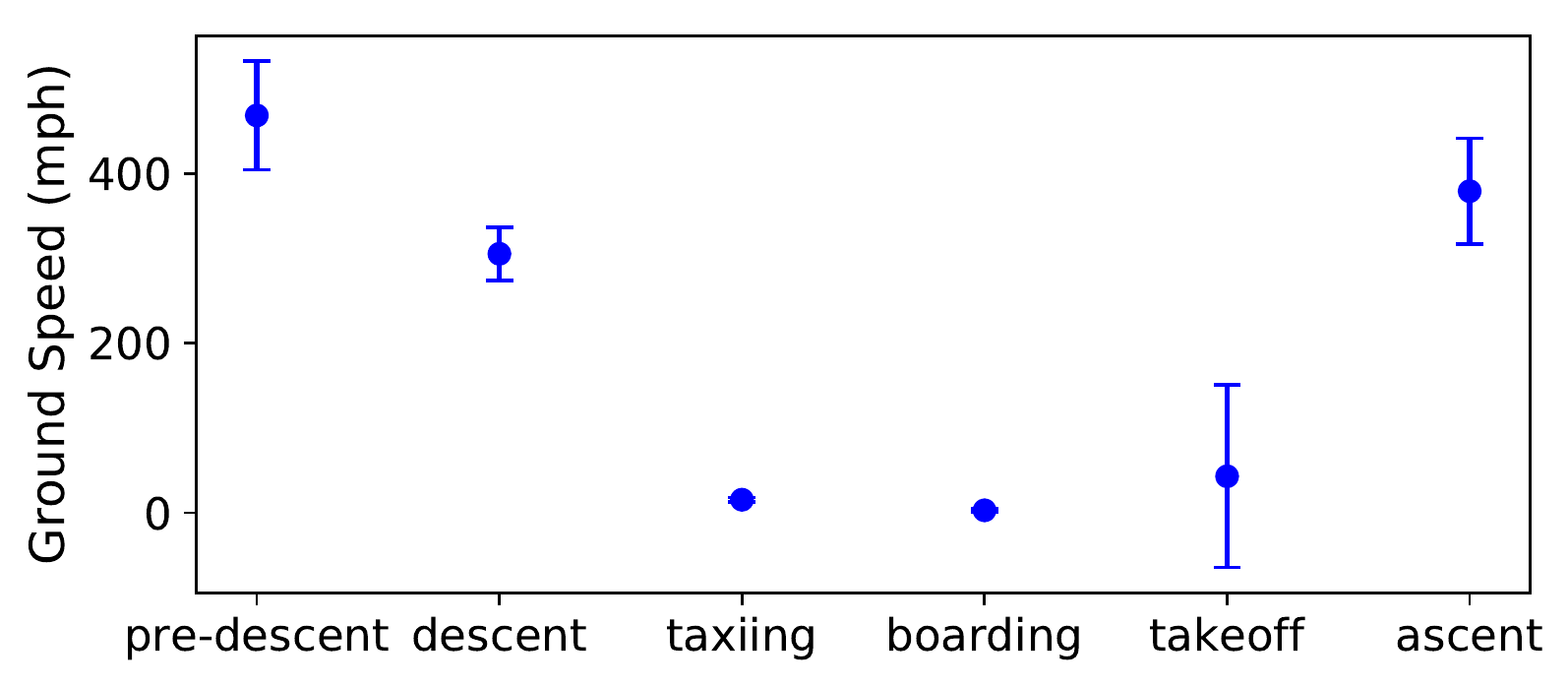} 
    \caption{Average altitude change (top) and ground speed (bottom) of each segment of the top motif  (J, G, A H, D, E) for all instances in the dataset. Error bars display one standard deviation.}
    \vspace{-4mm}
    \label{gspeed}
\end{figure}

Labeling each state from A through J, MASA identifies the top motif as $[J, G, A, H, D, E]$ (Figure \ref{altitude}). Plotting this motif against the airplane's altitude in Figure \ref{altitude}, we see that the motif encompasses the landing stages of each flight and the take-off stages of the next flight. This motif is extremely consistent, occurring almost every time the plane lands for one flight then takes off for the next (since we do not care about the length of each segment, the layover time does not effect the presence of this motif). We further characterize the plane's behavior by finding the average altitude change and ground speed for each of the six segments in the motif across all instances of the motif in the dataset. Using the velocities reported in Figure \ref{gspeed}, we can interpret each stage of the identified motif as:
\begin{itemize}
    \item \textbf{Pre-descent}: Equilibrium altitude, high ground speed.
    \item \textbf{Descent}: Slower ground speed with negative velocity indicating decrease in altitude.
    \item \textbf{Taxiing}: No vertical velocity and very slow ground speed.
    \item \textbf{Boarding}: Plane at rest. 
    \item \textbf{Takeoff}: Increased positive vertical velocity and ground speed. 
    \item \textbf{Ascent}: High positive vertical velocity and ground speed indicating ascent.
\end{itemize}

We note that MASA discovered this motif in a fully unsupervised manner. As shown, it was able to isolate an interesting repeated heterogeneous sequence of behaviors found across the time series. This has many practical benefits, as it can be used to auto-label the data with the different ``stages'' of a flight. Discovering motifs allows us to identify how these stages progress, and also lets us better label these stages when there is noise in the readings (due to turbulence, faulty sensors, or exogenous factors such as the weather).

\section{Conclusion and Future Work}
In this paper, we have developed a novel method for discovering motifs in time series data. Our approach, Motif-Aware State Assignment (MASA), leverages these motifs to better characterize and assign states, discovering repeated segments and relevant trends. %MASA is able to scale to very large time series, leveraging a high-performance alternating maximization algorithm that we have derived.
The promising results on both the synthetic experiments and case studies imply many potential directions for research. We leave for future work the analysis of MASA with different underlying likelihood models, rather than only TICC, the model we used in this paper. Furthermore, %MASA does not incorporate the lengths of the segments in the motifs, a simplifying assumption but one that may limit the accuracy in certain scenarios.
extending MASA to account for segment length, similar to hidden semi-Markov models, would open up this work to new applications and use cases.

%
% The acknowledgments section is defined using the "acks" environment (and NOT an unnumbered section). This ensures
% the proper identification of the section in the article metadata, and the consistent spelling of the heading.
% \begin{acks}
% \end{acks}

%
% The next two lines define the bibliography style to be used, and the bibliography file.
\bibliographystyle{ACM-Reference-Format}
\bibliography{sample-base}

% 
% If your work has an appendix, this is the place to put it.
\pagebreak
\appendix
\section{Implementation Details for Motif Candidate Set Generation}
\label{candidateAppendix}

Here we give a more detailed algorithm and implementation details for generating a candidate motif set. Each candidate motif $m$ is a list of states. Given our candidate set of motifs generated from the suffix array, we seek output a non-redundant set of relevant, significant motif candidates. 

The null model keeps track of a set $\mathcal{D}$, which contains repeated subsequences that have already been accepted as candidates. Initially, we set the null model to automatically know about repeats of size 2 that occurred in $\bbS$, which the null model assumes appear with probability according to their empirical frequency.

\begin{algorithm}[t!]
    
    \Fn{ReplaceKnownCands ($m$)}{
        Sort $\mathcal{D}$ by decreasing length, breaking ties by decreasing empirical probability.\;
        $m' \leftarrow m$
        \ForEach{$d\in\mathcal{D}$ } {
            $m' \leftarrow $ Replace non-overlapping instances of $d$ in $m'$ from left to right with $\phi(d)$
        \Return $m'$
        }}
    \caption{Replacement of Known Candidates}
    \label{replaceknowncands}
    \end{algorithm}   

    \begin{algorithm}[t!]
        \KwData{$\mathcal{C}$: the set of candidate motifs}
        Sort $\mathcal{C}$ by decreasing length.\;
        $\mathcal{D} \leftarrow$ The set of repeats of size 2.\;
        $\mathcal{A} \rightarrow \{\}$
        \ForEach{$m \in \mathcal{C}$} {
                $m' \leftarrow $ \texttt{ReplaceKnownCands}($m$)\;
            $p_\emptyset \leftarrow \prod_{m'_i \in m'} P_\emptyset(m'_i)$\;
            \If{$P\bigg(
                \mathcal{B}(|\bbS'|, p_{\emptyset} \geq N_m \bigg) \leq \frac{\alpha}{|\mathcal{C}|},$}{
                    Add $m$ to $\mathcal{A}$\;
                    Add $m$ to $\mathcal{D}$\;
                }
        }
        \Return $\mathcal{A}$.
        \caption{Motif Candidate Set Generation}
        \label{motifgenerationalg}
        \end{algorithm}

For a subsequence $d$, let the non-overlapping count of $d$ in $\bbS$ be $N_d$.  Denote $\phi(d)$ as a dummy ``state" which occurs with probability $P_{\emptyset}(\phi(d)) = \frac{N_d}{|\bbS|}$ according to the null model. Given a candidate motif $m$, we define the sub-routine \texttt{ReplaceKnownCands} which returns a new $m'$ which replaces known subsequences in $m$ with their fake states (Algorithm \ref{replaceknowncands}). The null model then evaluates the probability of the new $m'$ under the assumption that each state $m'_i$ (which might be real or ``dummy'') occurs independently according to its empirical probability.

We then reject candidates who do not have a high enough threshold $\alpha$. We use a Bonferroni corrected $\alpha=0.001$ normalized by the number of motif candidates being evaluated~\cite{napierala2012bonferroni}. The full algorithm is in Algorithm \ref{motifgenerationalg}.

\section{Reproducibility Details for our Experimental Evaluation}

We performed a total of 2 experiments (synthetic and cycling data) and 2 case studies (automobile and airplane data). For the two experiments, we pick $K$ (our number of states) to be the ground truth number of states in the constructed dataset. We then chose $\beta$ via BIC. For the case studies, we performed BIC to identify the cluster number. In both the cycling and airplane dataset, we performed PCA to reduce the dimensionality of each measurement to 10 and 13 features respectively. Scripts to run the experiments as well as the packaged results can be found in our linked source code. 

While details of the hyperparameters can be found in our packaged software (see Section), below is a table of the hyperparameters used for each experiment. 
\begin{table}[t!]
    \centering
\bgroup
\def\arraystretch{1}
\setlength\tabcolsep{3pt}
\begin{tabular}
{|r|| c c c |}
\hline
Experiment  & $K$ & $\beta$ & $\gamma$\\ \hline  \hline
Synthetic  & $10$ & $25$& $0.8$\\ \hline
Cycling &$ 8$  &$ 100$  & 0.8  \\ \hline
Airplane &$ 10$  &$ 50$  &$ 0.8$ \\ \hline
Automobile  &$ 8$  &$ 50$  &$ 0.6$  \\ \hline
\end{tabular}
\egroup
\vspace{2mm}
\caption{Hyperparameters for experiments}
\vspace{-5mm}
\end{table}

\end{document}